\documentclass[11pt]{article}
\usepackage{coling2020}
\usepackage{times}
\usepackage{url}
\usepackage{latexsym}

\usepackage{microtype}

\colingfinalcopy

\usepackage[utf8]{inputenc}
\usepackage{enumitem}
\usepackage{subfig}
\usepackage{graphicx}
\usepackage{caption}
\usepackage{booktabs}
\usepackage{float}
\usepackage{amssymb}
\usepackage{mathtools}
\usepackage{amsmath}
\usepackage{multirow}
\usepackage{tabu}
\usepackage{pifont}
\usepackage{longtable}
\usepackage{xspace}

\usepackage[capitalise,noabbrev]{cleveref}
\usepackage{todonotes}
\usepackage[english]{babel}
\usepackage[autostyle]{csquotes}
\MakeOuterQuote{"}

\graphicspath{{./figures/}}
\newcommand{\sota}{state-of-the-art\xspace}

\newcommand{\bertuni}{\textsc{BertUni}\xspace}
\newcommand{\bertbi}{\textsc{BertBi}\xspace}
\newcommand{\bertbiss}{\textsc{BertBiSS}\xspace}
\newcommand{\bertbitf}{\textsc{BertBi-TF}\xspace}
\newcommand{\berttri}{\textsc{BertTri}\xspace}
\newcommand{\berttriss}{\textsc{BertTriSS}\xspace}
\newcommand{\hb}{$\uparrow$\xspace} %
\newcommand{\lb}{$\downarrow$\xspace} %
\newcommand{\bsp}[1]{%
  \bgroup
  \itshape
  #1%
  \egroup%
}
\newcommand{\ts}{\thinspace} %

\title{Evaluation Discrepancy Discovery: A Sentence Compression Case-study}

\author{
  Yevgeniy Puzikov\thanks{~~Research done during an
    internship at Bloomberg L.P., London, United Kingdom.} \\
  Ubiquitous Knowledge Processing Lab (UKP Lab), \\
  Department of Computer Science, Technical University of Darmstadt \\
  \url{https://www.ukp.tu-darmstadt.de}
}

\date{}

\begin{document}
\maketitle

\begin{abstract}
  Reliable evaluation protocols are of utmost importance for
  reproducible NLP research. In this work, we show that sometimes
  neither metric nor conventional human evaluation is sufficient to
  draw conclusions about system performance. Using sentence
  compression as an example task, we demonstrate how a system can game
  a well-established dataset to achieve state-of-the-art results. In
  contrast with the results reported in previous work that
  \emph{showed correlation} between human judgements and metric
  scores, our manual analysis of state-of-the-art system outputs
  demonstrates that high metric scores \emph{may only indicate a
    better fit to the data}, but not better outputs, as perceived by
  humans. The prediction and error analysis files are publicly
  released.
  \footnote{\url{https://github.com/UKPLab/arxiv2021-evaluation-discrepancy-nsc}}
\end{abstract}

\section{Introduction}\label{sec:intro}
\subsection{Task description}
Sentence compression is a Natural Language Processing (NLP) task in
which a system produces a concise summary of a given sentence, while
preserving the grammaticality and the important content of the
original input. Both
abstractive~\cite{cohn2008sentence,rush2015neural} and
extractive~\cite{filippova2013overcoming,filippova-etal-2015-sentence,wang-etal-2017-syntax,zhao2018language-sentence-compression}
approaches have been proposed to tackle this problem. Most researchers
have focused on the extractive methods, which treat this as a
deletion-based task where each compression is a subsequence of tokens
from its original sentence (\Cref{fig:nsc_data_specification}).

\begin{figure}[h]
  \centering  
  \subfloat[Input sentence]{
    \noindent
    \parbox{0.45\textwidth}{%
      \bsp{Dickinson, who competed in triple jump at the 1936 Berlin
        Games, was also a bronze medalist in both the long jump and
        triple jump at the 1938 Empire Games.}}%
    }
    \\
    \subfloat[Reference compression]{
      \noindent
      \parbox{0.45\textwidth}{%
        \bsp{Dickinson competed in triple jump at the 1936 Berlin
          Games.}}%
    }
    \\
    \subfloat[Another possible compression]{
      \noindent
      \parbox{0.45\textwidth}{%
        \bsp{Dickinson was a bronze medalist in the long jump and
          triple jump at the 1938 Empire Games.}  }%
    }
    
    \caption[Sentence Compression Example]{Sentence compression
      example from the Google Dataset: an input sentence and a
      reference compression. The compression candidate at the bottom
      is also a valid one, but would score low, because the n-gram
      overlap with the reference is small.}
  \label{fig:nsc_data_specification}
\end{figure}

In the past few years several novel methods have been proposed to
tackle the task of sentence compression. Most of these methods have
been evaluated using the Google
Dataset~\cite{filippova2013overcoming} or its derivatives. Most
authors present approaches that show better metric scores; a few of
them also describe human evaluation experiments and show that the
proposed methods outperform previous work. However, there has been a
serious lack of analysis done on actual model predictions. In this
work we show that metric scores obtained on the Google Dataset might
be misleading; a closer look at model predictions reveals a
considerable amount of noise, which renders the trained model
predictions ungrammatical. Another problem is that valid system
outputs which do not match the references are severely penalized. For
example, a plausible compression for the introductory example we used
above would be \bsp{Dickinson was a bronze medalist in the long jump
  and triple jump at the 1938 Empire Games.} However, with the
established evaluation protocol, this compression would score very low
because of the insignificant token overlap with the reference. We
showed that evaluating a system on the Google Dataset is tricky, and
even human evaluation done in the previous years could not detect the
issues described in this chapter.

To summarize, our contributions in this study are:
\begin{itemize}
  \item We introduce a simple method of sentence compression that established new \sota results, as measured by common metrics.
  \item We design an experiment with a contrived system which achieved even higher scores, but produced less grammatical and less informative outputs.
    \item We show that this discrepancy may be attributed to the noise in the dataset.
\end{itemize}

\section{Data Analysis}\label{sec:nsc_data_analysis}
In our experiments we use the Google Dataset introduced
by~\newcite{filippova2013overcoming}~\footnote{\url{https://bit.ly/2ZvTK9z}}
This dataset was constructed automatically by collecting English news
articles from the Internet, treating the first sentence of each
article as an uncompressed sentence and creating its extractive
compression using a set of heuristics and the headline of the
article. The dataset contains 200\ts000 training and 10\ts000
evaluation instances; the first 1\ts000 data points from the latter
are commonly used as a test set and the remaining 9\ts000 as a
development set.

Exploratory data analysis showed that the distribution of the training
data is highly skewed, which is not surprising though, given the
nature of the data.

\begin{figure*}[t]
  \centering
  \newcommand\x{0.45}
  \subfloat[Sentence length]{%
    \includegraphics[width=\x\textwidth]{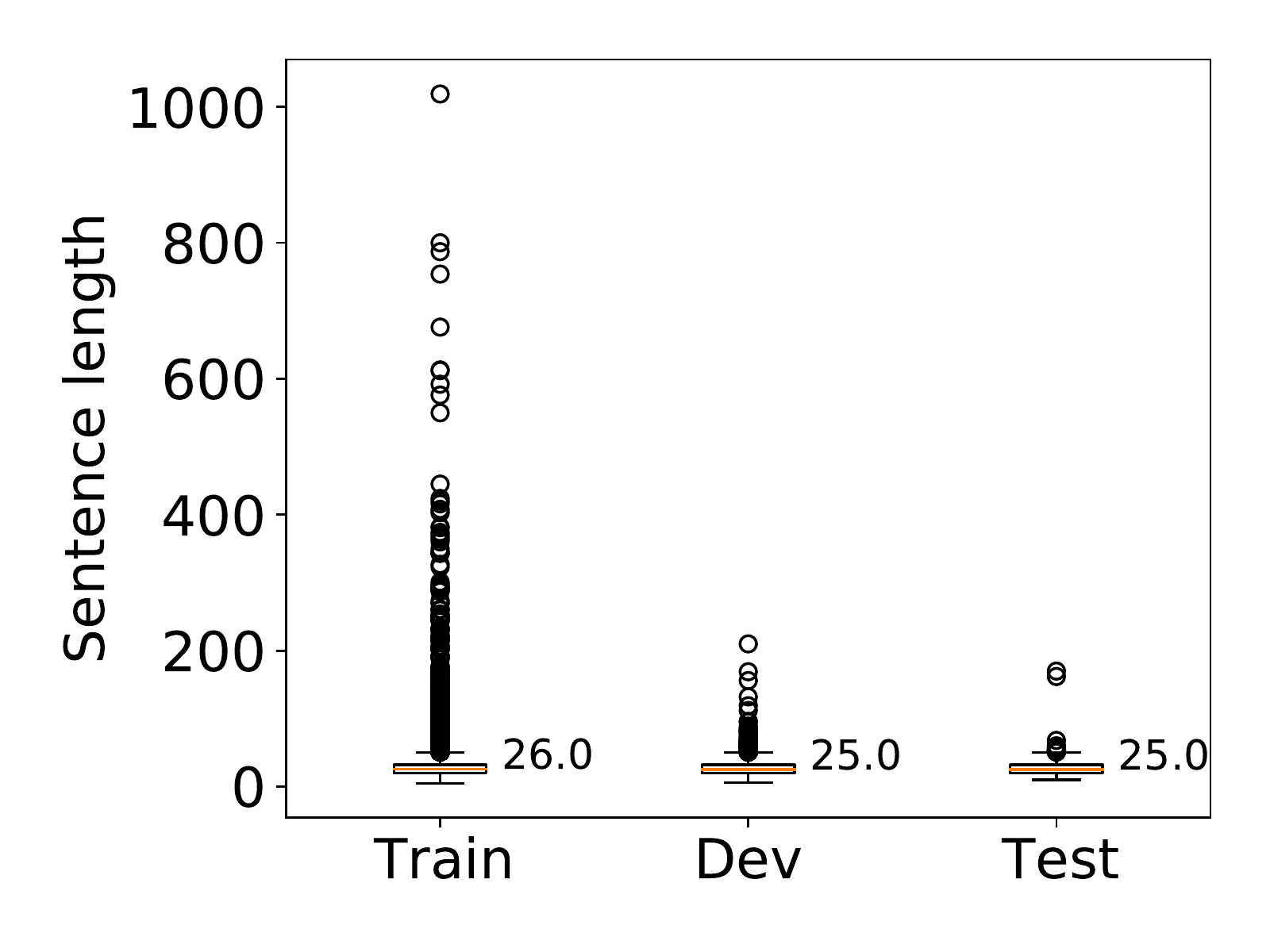}
    \label{fig:nsc_data_analysis_snt}}
  \subfloat[Reference length]{
    \includegraphics[width=\x\textwidth]{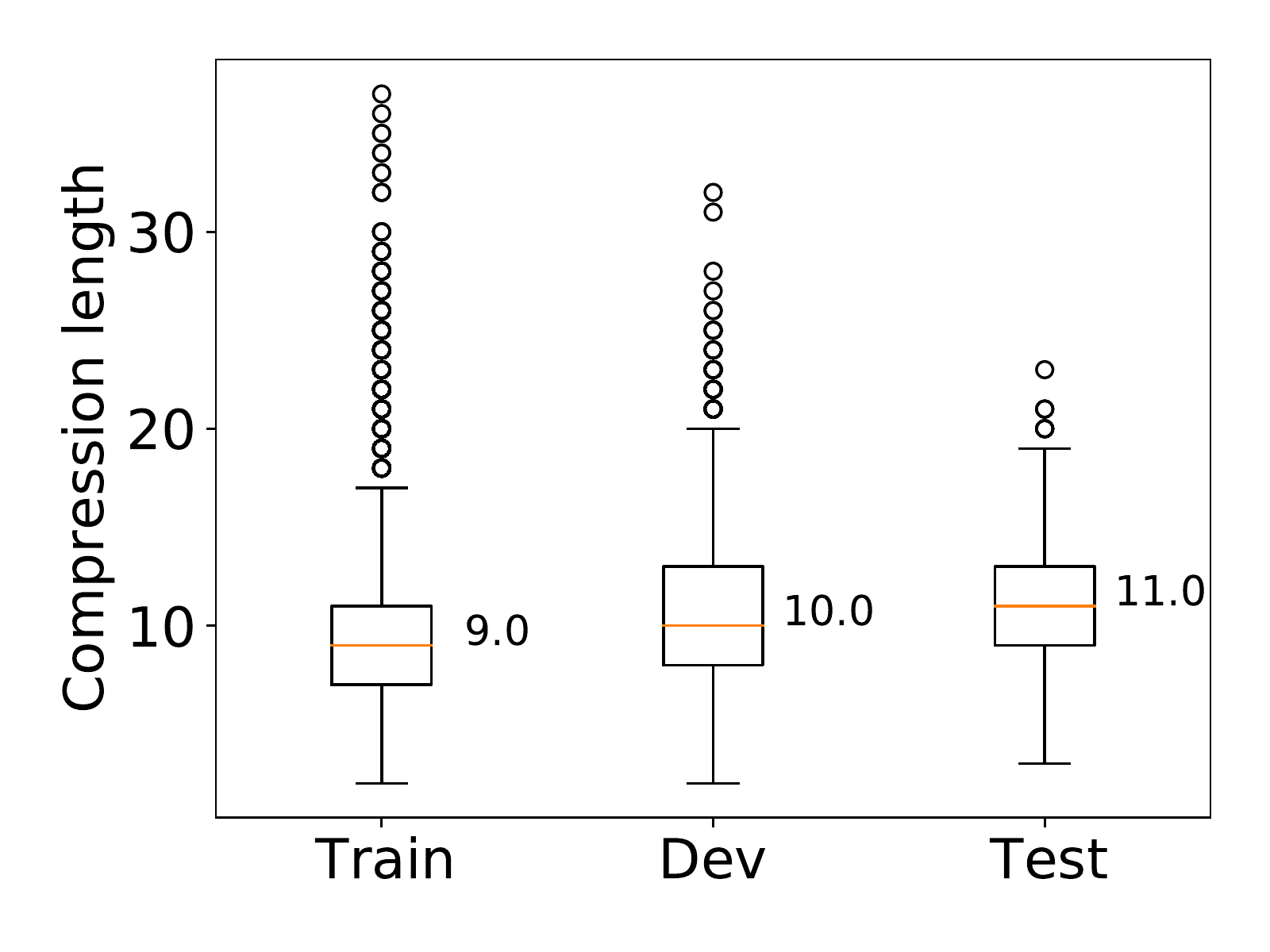}
    \label{fig:nsc_data_analysis_cmp}}
  \qquad
  \subfloat[Token length]{
    \includegraphics[width=\x\textwidth]{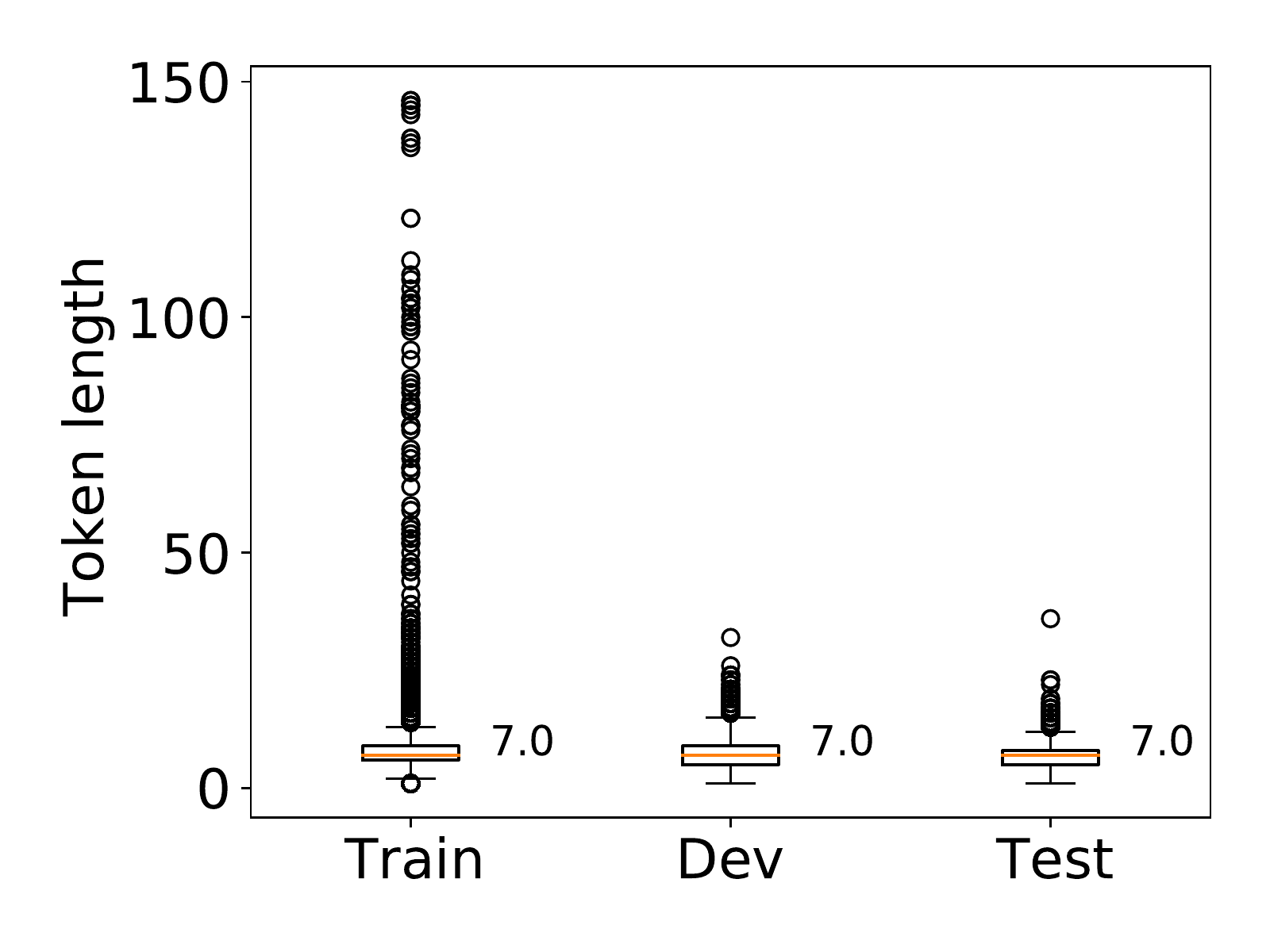}
    \label{fig:nsc_data_analysis_tok}}
  \subfloat[Compression ratio (CR) values]{
    \includegraphics[width=\x\textwidth]{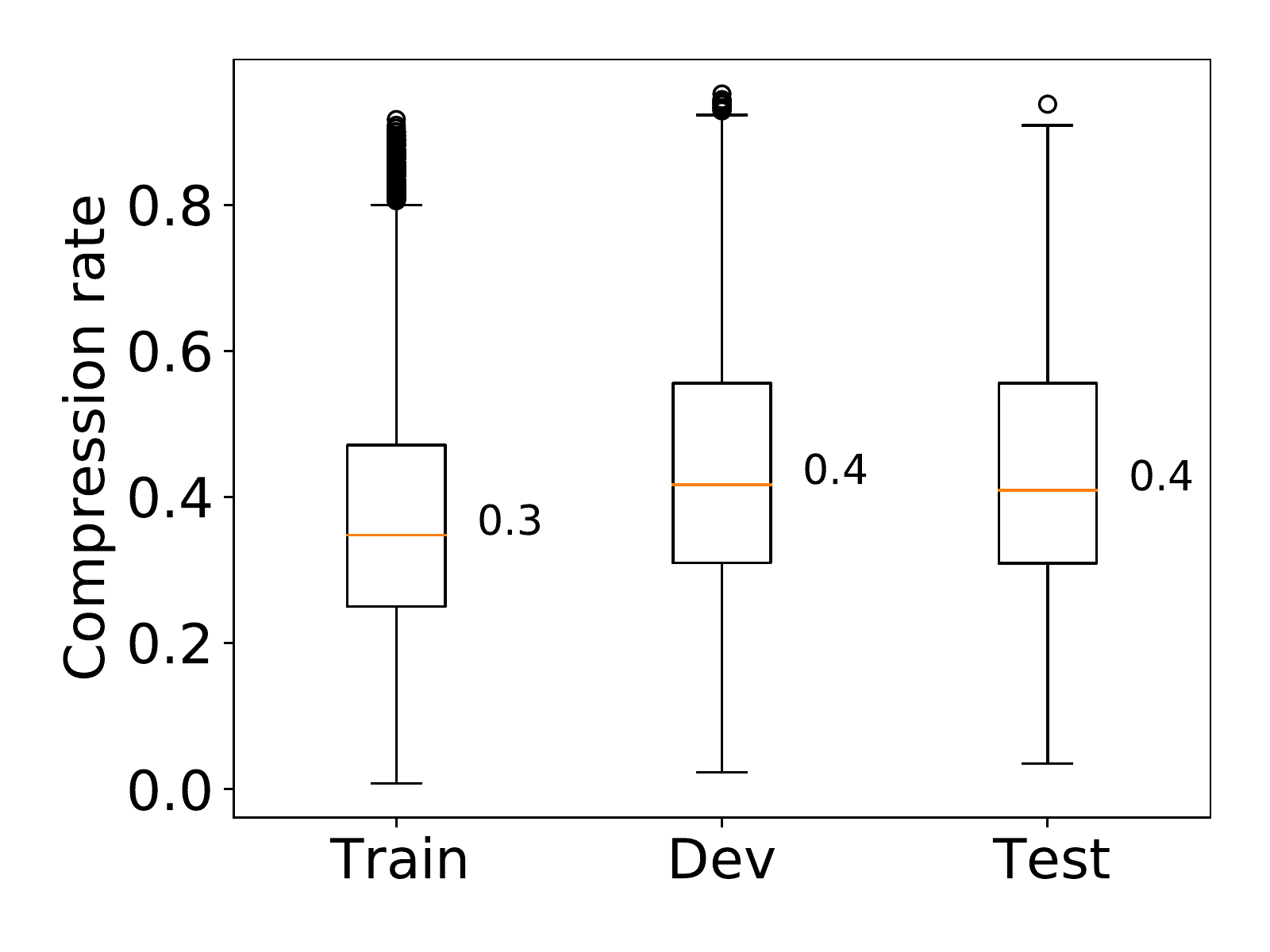}
    \label{fig:nsc_data_analysis_cr}}
  \caption[Google Dataset Analysis]{Data analysis of the Google
    Dataset: length distributions of sentences, ground-truth
    compressions and tokens, and the distribution of compression ratio
    values. The numbers to the right of each box denote median
    values.}
  \label{fig:nsc_data_analysis_train}
\end{figure*}

In order to remove outliers and fit the computation budget, we removed
instances which contained sentences longer than 50 tokens and
compressions longer than 17 tokens. We also removed examples with
tokens longer than 15 characters, since those in most cases denoted
website links. Finally, we excluded cases with a compression ratio of
more than 0.85~--- those rare cases in most cases were too long to
qualify as compressions. Evaluation on the development and test sets
was done without any data filtering.

\section{BERT-based Sentence Compression}\label{sec:nsc_our_approach_uni}

Most modern deletion-based compression systems adopt either a
tree-pruning, or a sequence labeling approach.  The former uses
syntactic information to navigate over a syntactic tree of a sentence
and decide which parts of it to
remove~\cite{knight2000statistics,mcdonald-2006-discriminative,filippova2013overcoming}. With
the advent of sequence-to-sequence models it became possible to skip
the syntactic parsing step and solve the task directly, by processing
a sentence one token at a time and making binary decisions as to
whether to keep a token or delete it~\cite{filippova-etal-2015-sentence,wang-etal-2017-syntax,zhao2018language-sentence-compression,kamigaito2020slahan}. The
advantages of such approaches include a lesser chance of introducing
error propagation from incorrect parsing decisions, as well as higher
training and inference speed.

For a long time the space of sequence-to-sequence models has been
dominated by different variants of Recurrent Neural Networks
(RNN)~\cite{rumelhart1986learning}. However, a more recent Transformer
architecture~\cite{vaswani-2017-transformer} has shown very promising
results in many NLP tasks. Given the success of Bidirectional Encoder
Representations from Transformers (BERT)~\cite{devlin-2019-bert}, and the
fact that there has been no empirical evaluation of its performance in
sentence compression, we decided to fill this gap and find out how
well BERT-based models would cope with the task.

\begin{figure}
  \centering
  \includegraphics[scale=0.7]{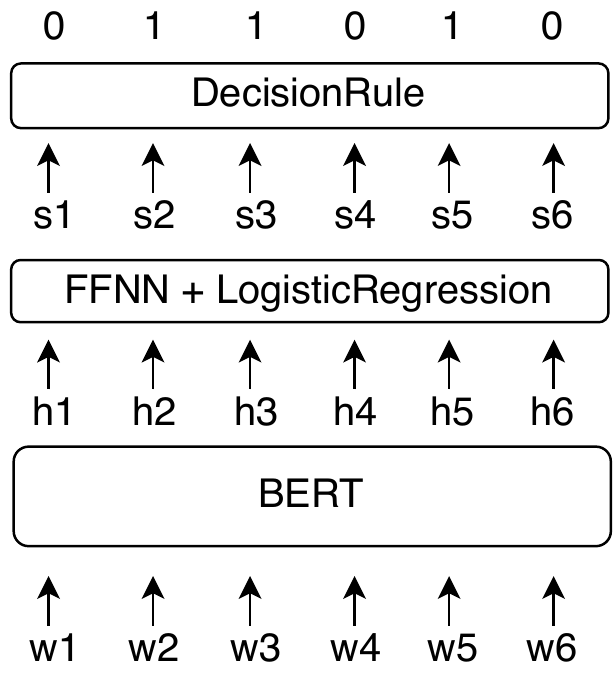}
  \caption{Schematic of the \bertuni architecture.}
  \label{fig:nsc_bertuni}
\end{figure}

We used pretrained BERT-base-uncased model
weights~\footnote{\url{https://huggingface.co/bert-base-uncased}}
provided by the HuggingFace library~\cite{wolf2020transformers}, and
implemented a simple \bertuni model which encodes the source sentence
$ S = \{w_1, w_2, \dots w_n\}$ and produces a sequence of vectors
$ V = \{v_1, v_2, \dots v_n\}, v_i \in \mathbb{R}^h$. Each vector is
fed into a dense layer with a logistic function as a non-linear
function to produce a score $s_i \in [0,1]$
(\Cref{fig:nsc_bertuni}). If $s_i \geq 0.5$, the model output is 1
(and 0, otherwise).

A simple decision rule was used to make a binary prediction: 

\begin{equation*}
  \label{decision_rule}
  decision =
  \begin{cases}
    \text{1}, &   \text{if } \mathit{s} \geq 0.5,\\
    \text{0},  &   \text{otherwise}.
  \end{cases}
\end{equation*}

\section{Experiments}\label{sec:nsc_eval}

\subsection{Automatic Metric Evaluation}\label{sec:nsc_eval_auto}
For automatic evaluation of sentence compression systems, most
researchers follow~\newcite{clarke2006models,filippova2013overcoming}
and use the following two metrics:

\begin{itemize}
\item \textbf{F1-score}: harmonic mean of the recall and
    precision in terms of tokens kept in the target and the generated
    compressions.
\item \textbf{Compression ratio (CR)}: the length of the compression divided over the original sentence length.
\end{itemize}

The former metric shows how close the model outputs are to the
references. The latter one is supposed to measure the compression
effectiveness.

To make our results comparable to previous work, in our experiments we
followed the same convention. However, we would like to note that
\emph{measuring CR in sentence compression might be redundant} for
several reasons. The first reason comes from the fact that data-driven
sentence compressors are likely to produce outputs with a compression
ratio most commonly seen in the training data references. In other
words, CR is less a property of a system and more a characteristic of
the dataset. This is supported by the fact that most models reported
in the literature have the same compression ratio (in the range of
0.38--0.43, see~\Cref{tab:nsc_eval_metric_bertuni}).

Secondly, it is not even clear how to treat compression ratio values:
is a CR of 0.4 better than a CR of 0.5? Intuitively, yes, because it
means a more concise compression. However, the compression is really
better only if it retained more valuable information from the
source. On the other side, defining the notion of
informativeness/importance in sentence compression (and document
summarization, in general) is an open problem and currently is not
measured automatically. This means that the CR metric is a very
one-sided proxy, too crude to be used for real automatic evaluation
without balancing it with some recall-oriented metric.

To put the evaluation of our approach into better context, we compare
it with the following systems. All systems predict a sequence of
binary labels which decide which tokens to keep or remove from the
input sentence.

\paragraph{LSTM.} \newcite{filippova-etal-2015-sentence} use a
three-layer uni-directional Long Short-term Memory (LSTM)
network~\cite{hochreiter-1997-lstm} and pretrained
\textsc{word2vec}~\cite{mikolov-2013-distributed} embeddings as input
representations. For comparison, we use the results for the best
configuration reported in the paper (\textsc{LSTM-PAR-PRES}). This
system parses the input sentence into a dependency tree, encodes the
tree structure and passes the aggregated feature representations to
the decoder LSTM. Unlike our approach, this system relies on beam
search at inference time.

\paragraph{BiLSTM.} \newcite{wang-etal-2017-syntax} build upon \textsc{LSTM}
approach, but introduces several modifications. It employs a
bi-directional LSTM encoder and enriches the feature representation
with syntactic context. In addition, it uses Integer Linear
Programming (ILP) methods to enforce explicit constraints on the
syntactic structure and sentence length of the output.

\paragraph{Evaluator-LM.}
\newcite{zhao2018language-sentence-compression} uses a bi-directional
RNN to encode the input sentence and predict a binary label for each
input token. In addition to token embeddings, the network uses vector
representations of part-of-speech (POS) tags and dependency
relations. The system is trained using the \textsc{REINFORCE}
algorithm~\cite{williams1992reinforce}, the reward signal comes from a
pretrained syntax-based language model (LM).

\paragraph{SLAHAN.} \newcite{kamigaito2020slahan} propose a modular
sequence-to-sequence model that consists of several components. The
system encodes a sequence of tokens using a combination of pretrained
embeddings (Glove~\cite{pennington2014glove},
ELMO~\cite{peters2018deep}, BERT) and parses the input into a
dependency graph. Three attention modules are employed to encode the
relations in the graph, their weighted sum is passed to a selective
gate. The output of the latter forms an input to a LSTM decoder.

Despite its simplicity, the proposed BERT-based approach achieved very
competitive scores (\Cref{tab:nsc_eval_metric_bertuni}).
\begin{table}[]
  \centering
  \small
    \begin{tabular}{@{}lrr@{}}
      \toprule
      \textbf{Model}          & \textbf{F1}\hb & \textbf{CR}\lb \\
      \midrule
      \textsc{Evaluator-LM} & 0.851 & 0.39 \\
      \textsc{BiLSTM}  & 0.800       & 0.43   \\
      \textsc{LSTM} & 0.820   & \textbf{0.38}  \\
      \textsc{SLAHAN} & 0.855 & 0.407 \\ 
      \bertuni & \textbf{0.857} $\pm 0.002$  & 0.413 $\pm 0.004$ \\
      \midrule
      \bertuni (dev) & \textbf{0.860} $\pm 0.001$       & 0.418 $\pm 0.004$   \\
      \bottomrule
    \end{tabular}%
  \caption{The performance of the \bertuni model on the test portion
    of the Google Dataset, compared to recent approaches. The last row
    shows \bertuni's performance on the development set.}
  \label{tab:nsc_eval_metric_bertuni}
\end{table}

Comparing single performance scores (and not score distributions) of
neural approaches is meaningless, because training neural models is
non-deterministic in many aspects and depends on random weight
initialization, random shuffling of the training data for each epoch,
applying random dropout masks~\cite{reimers-2017-reporting}. This
makes it hard to compare the scores reported in previous works and our
approach. To facilitate a fair comparison with future systems, we
report the mean and standard deviation of the \bertuni scores averaged
across ten runs with different random seeds.

In order to understand where \bertuni fails and what we could
potentially improve upon, we conducted manual error analysis of its
predictions.

\subsection{Error analysis}\label{sec:nsc_erran_uni}

The purpose of error analysis is to find weak spots of a system, from
the point of view of human evaluation. In sentence compression,
previous work typically analyzed system predictions of the first 200
sentences of the test set, using a 5-point Likert scale to assess
annotators' opinions of the compressions' \emph{readability} and
\emph{informativeness}~\cite{filippova-etal-2015-sentence}.

Since error analysis is used for further system improvement and test
sets should be used only for final evaluation, we perform error
analysis on the development set. In order to do that, we retrieved
\bertuni's predictions on the 200 dev set sentences which received the
lowest F1 scores and manually examined them. Note that those are not
random samples; the reason why we chose worst predictions is because we know
that the system performed poorly on them.

As for the quality criteria, we had to make certain
adjustments. \newcite{filippova-etal-2015-sentence} mention that
readability \emph{covers the grammatical correctness,
  comprehensibility and fluency of the output}, while informativeness
measures \emph{the amount of important content preserved in the
  compression}. In our opinion, merging several criteria into one
synthetic index is a bad idea, because annotators can't easily decide
on the exact facet of evaluation. Given that there already exists a
problem of distinguishing fluency and grammaticality, adding both of
them to assess readability seems to be a bad design decision. The
problem is aggravated by the fact that readability as a text quality
criterion is already used by NLP researchers for estimating the
\emph{text complexity} from a reader's point of
view~\cite{vajjala2012improving,stajner2013readability,venturi2015nlp,de2016all}. This
made us conclude that readability is another overloaded
criterion. Instead, we chose \emph{grammaticality} as the first
quality criterion.

We manually analyzed \bertuni predictions on the 200 aforementioned
samples, trying to identify common error patterns. The results are
presented below.

\begin{figure}[t]
  \centering
  \subfloat[\bertuni errors]{
    \includegraphics[width=0.3\textwidth]{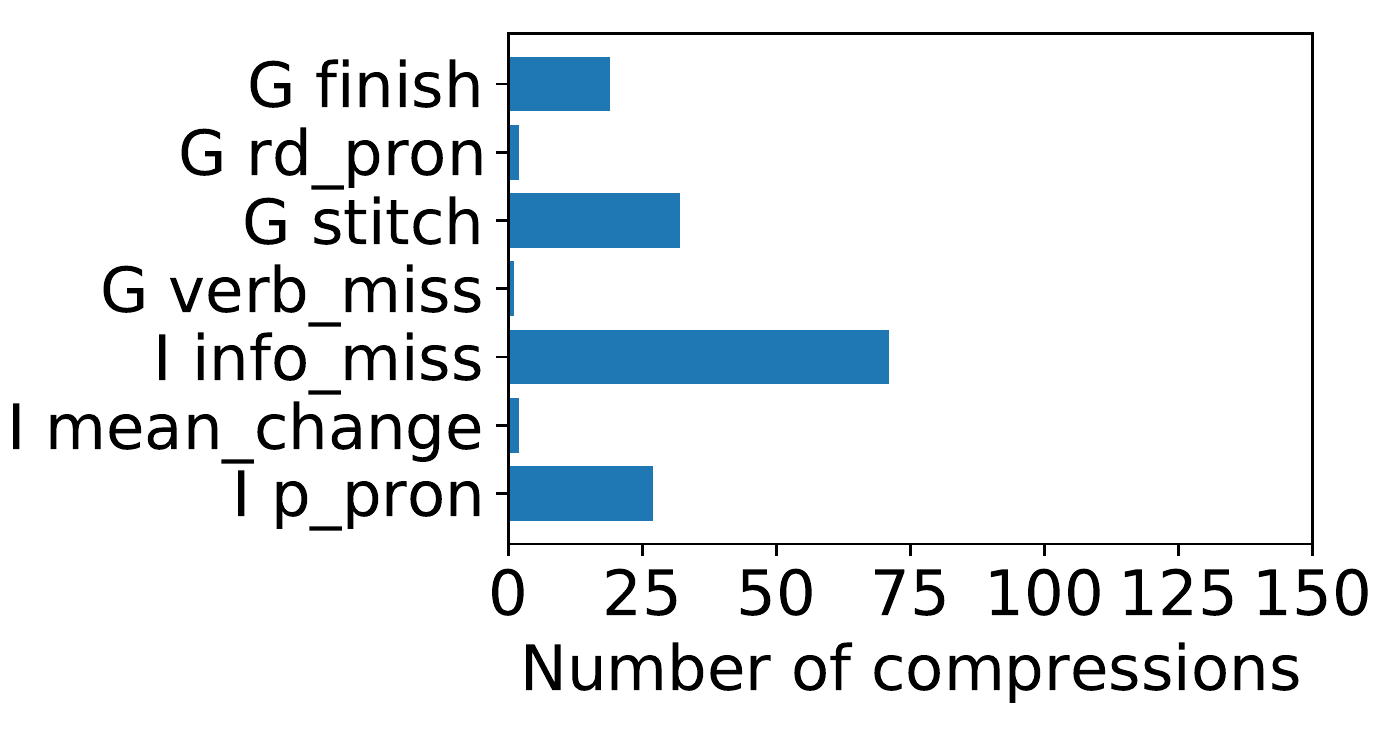}
    \label{fig:nsc_erran_uni}
  }
  \subfloat[\bertbitf errors]{
    \includegraphics[width=0.3\textwidth]{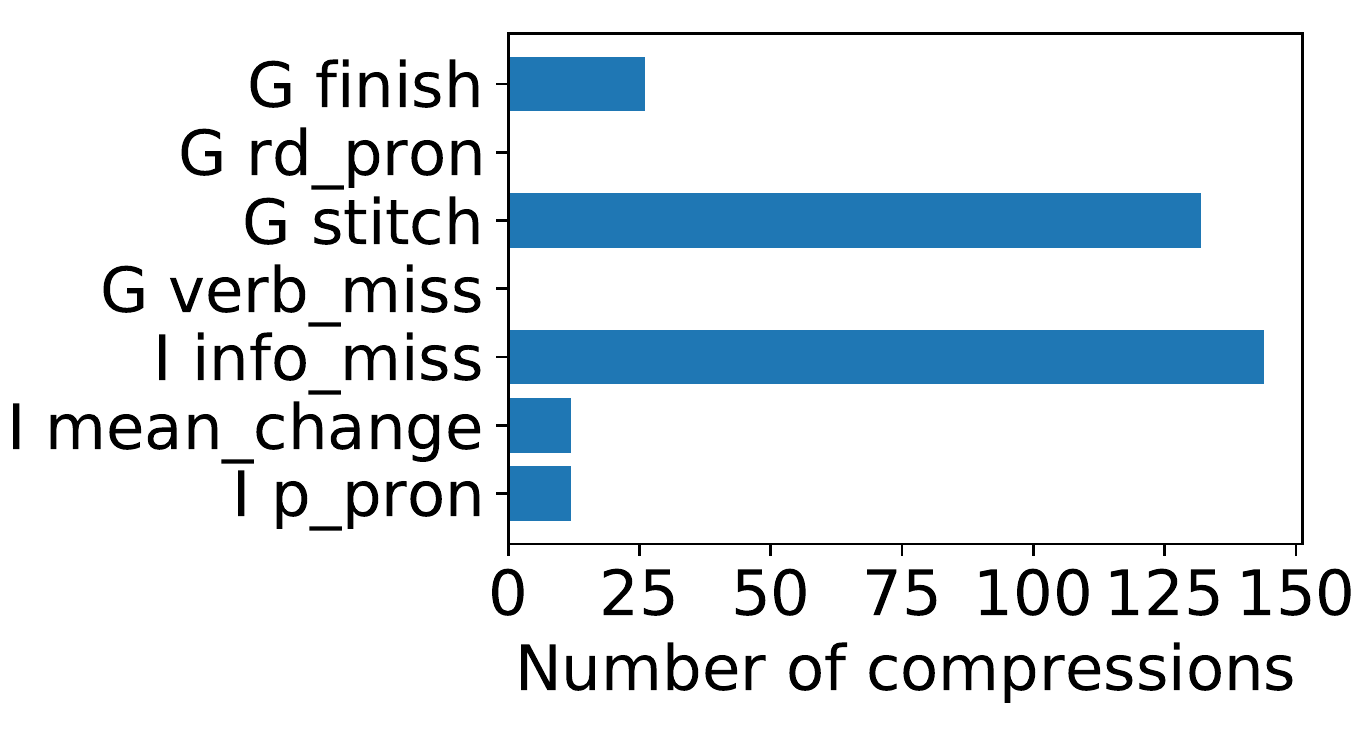}
    \label{fig:nsc_erran_bitf}
  }
  \subfloat[Ground-truth errors]{
    \includegraphics[width=0.3\textwidth]{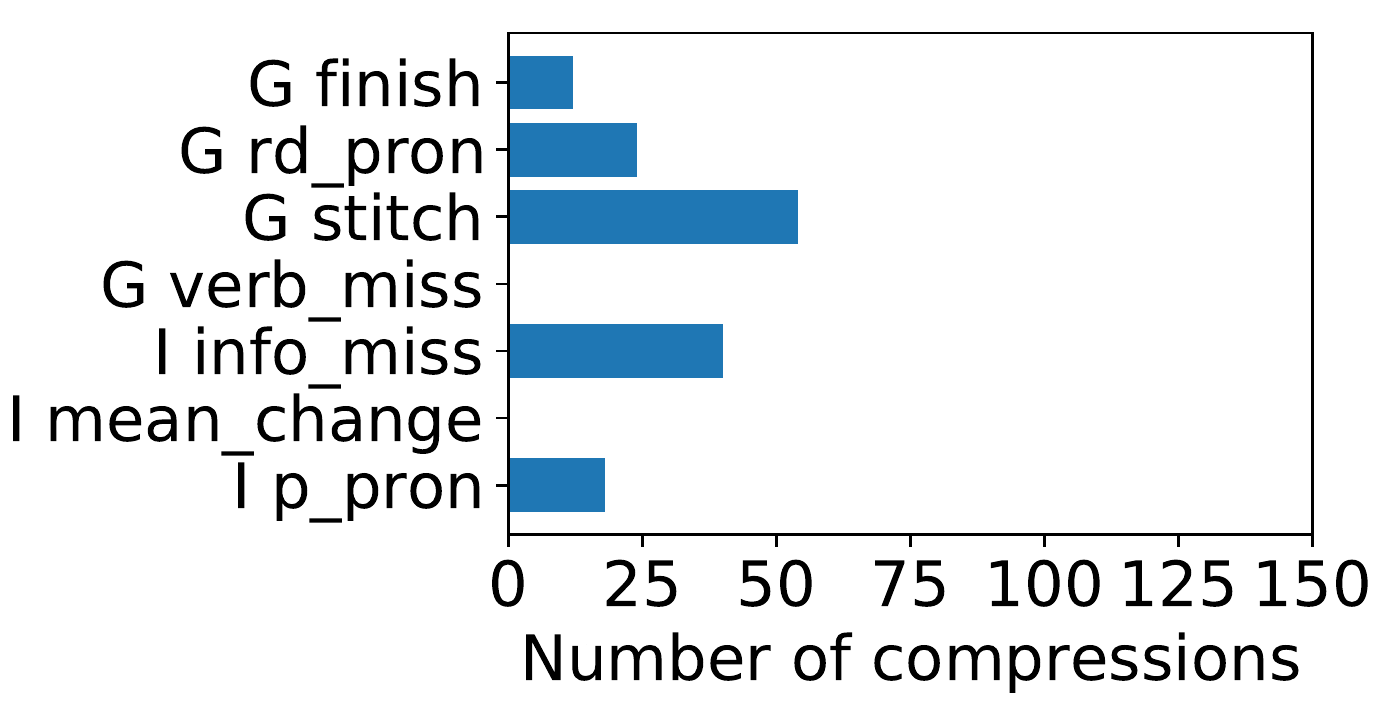}
    \label{fig:nsc_erran_tgt}
  }
  \caption{Number of errors made by the evaluated approaches on the 200
    development set instances where \bertuni achieved the lowest F1
    scores, as well as errors found in ground-truth compressions.
    Error types marked with \emph{G} are \emph{grammaticality} flaws; the
    remaining ones are errors of \emph{informativeness}.}
  \label{fig:nsc_erran}
\end{figure}

\paragraph{Grammaticality.} Out of 200 compressions, 146 (73 \%) were
deemed to be grammatical. The errors in the remaining instances have
been classified into several groups (marked with \emph{G}
in~\Cref{fig:nsc_erran_uni}).

Most of them were cases where grammatical clauses miss linking words,
are \emph{stitched} together, making the output
ungrammatical, as in the following compressions:
\begin{itemize}
  \item \bsp{I 'm said It 's not Kitty Pryde superhero is the leader of the X-Men .}
  \item \bsp{He first Postal Vote result can be announced before 10PM .}
\end{itemize}

Another large error category was \emph{finish}: the compression was
grammatical until the last retained token, where the sentence ended
abruptly, rendering the compression incomplete:
\begin{itemize}
  \item \bsp{Activision Blizzard has confirmed some new statistics for its games including .}
  \item \bsp{The South Sydney star had no case to .}
\end{itemize}

A few system outputs incorrectly started with a relative or
demonstrative pronoun. This happened when the system failed to retain
parts of main clause of the sentence (\emph{rd-pron}):
\begin{itemize}
  \item \bsp{That shows young people rapping while flashing cash and a handgun in a public park .}
\end{itemize}

Finally, one output missed a verb which was essential for ensuring
grammaticality (\emph{verb-miss}):
\begin{itemize}
  \item \bsp{People giant waves crash against the railway line and buildings at Dawlish .}
\end{itemize}

\paragraph{Informativeness.} Out of 200 compressions, 105 (52.5 \%) were
deemed to be informative, the errors in the remaining instances have
been classified into several groups (marked with \emph{I}
in~\Cref{fig:nsc_erran_uni}). Most of these erroneous cases were
compressions which missed certain information that was needed for
understanding the context (\emph{info-miss}). For example:
\begin{itemize}
  \item \bsp{Dolly Bindra filed a case .}
  \item \bsp{Mount Hope became the third largest city .}
\end{itemize}

A smaller, but still a large group of compressions started with
unresolved personal pronouns, which made it hard to understand the
subject (\emph{p-pron}):
\begin{itemize}
    \item \bsp{She hopes her album Britney Jean will inspire people .}
    \item \bsp{He should be allowed to work freely till proven guilty .}
\end{itemize}

In some cases, omitting the context caused a change in the meaning of
the sentence (\emph{mean-change}). For example:
\begin{itemize}
  \item \textbf{Reference}: \bsp{[...] Aleksandar Vucic [\dots] voiced
hope that Germany will give even stronger support to Serbia [\dots]}
  \item \textbf{System}: \bsp{Aleksandar spoke Germany will give
stronger support to Serbia .}
\end{itemize}

A large number of both grammatical and informative compressions did
not match references (\emph{I2}). Interestingly enough, in some cases the system
outputs were better then the references:
\begin{itemize}
  \item \textbf{Reference}: \bsp{We saw their two and raised to three.}
  \item \textbf{System}: \bsp{Newport beat Hartlepool 2 0 .}
  \vspace{5mm}
  \item \textbf{Reference}: \bsp{Who joins for the remainder of the season subject .}
  \item \textbf{System}: \bsp{Watford have announced the signing of Lucas Neill .}   
\end{itemize}

More examples of compression errors are provided in~\Cref{sec:appendix_erran_uni}.

\section{Evaluation Discrepancy}\label{sec:nsc_our_approach_bi}

When assessing the sentence compressions, we needed to compare system
outputs with references. Manual examination revealed that many
references themselves were flawed. This, in turn, meant that noise is
inherent to the Google Dataset, and metric-based improvements on this data
are misleading. To corroborate this claim, we conducted two
experiments: the first tested the capacity of a more accurate system
to ignore the noise and output compressions of better quality. In the
second, we verified whether the noise came from the ground-truth data
and attempted to quantify it.

At first, we decided to implement more complex models that could
potentially achieve better scores. We attempted to improve the
grammatical quality of \bertuni compressions by using the history of
model predictions for making more informed decisions.

We impelented and tested models that use BERT-encoded lastly-retained
tokens at each prediction step as an additional input to the model
(prediction history), similar to n-gram language models. As a history,
\bertbi and \berttri used one and two previously predicted tokens,
respectively. \bertbiss and \berttriss were the same as \bertbi and
\berttri, but used scheduled sampling training scheme to mitigate the
exposure bias issue~\cite{bengio-2015-scheduled}.

According to the metric evaluation results, none of the more complex
models outperformed \bertuni (\Cref{tab:nsc_eval_metric_variants}).

\begin{table}
  \centering
  \small
    \begin{tabular}{@{}lrr@{}}
      \toprule
      \textbf{Model}          & \textbf{F1}\hb & \textbf{CR}\lb \\
      \midrule
      \bertuni{} & \textbf{0.860} $\pm 0.001$       & 0.418 $\pm 0.004$       \\
      \bertbi{} & 0.849 $\pm 0.001$       & 0.423 $\pm 0.005$        \\
      \bertbiss{} & 0.840 $\pm 0.003$      & \textbf{0.370}  $\pm 0.005$       \\
      \berttri{} & 0.847 $\pm 0.002$      & 0.423 $\pm 0.007$        \\
      \berttriss{} & 0.843 $\pm 0.003$       & 0.382 $\pm 0.006$        \\
      \midrule
      \bertbitf{} & 0.901       & 0.423        \\
      \bottomrule
    \end{tabular}%
  \caption{BERT-based model variants' performance on the development
    set (mean and standard deviation across ten random seed
    values). \bertbitf was run only once, since it is a ``cheating''
    model that is not meant to be used in production.}
  \label{tab:nsc_eval_metric_variants}
\end{table}

We used an unrealistic scenario and artificially made it easier for
the model to make correct predictions. We trained a \bertbitf model
which builds upon \bertbi, but at prediction time for history instead
of model predictions uses ground-truth labels~\footnote{We call this
  model \bertbitf, since it builds upon \bertbi, but uses teacher
  forcing (\textsc{TF}) both at training and prediction time.}. The
development set result of \bertbitf was an F1 score of \textbf{0.901},
a 4-point improvement over \bertuni. We retrieved this model's
predictions for the same 200 dev set sentences used for the error
analysis of \bertuni outputs, and manually examined them. The usual
evaluation practice is to draw samples randomly, in order to not give
an advantage to any system and not to bias the evaluation. However, in
this work we approached the problem from a system-development
perspective and attempted to assess the comparative performance of the
approaches in the \emph{worst-case} scenario. If such a comparison is
biased, then only in favor of \bertbitf, because the drawn samples
were the worst ones for \bertuni, not \bertbitf. We view this as
sanity step, a regression test to ensure that the newer version of the
system performs at least as well as the baseline on the challenging
cases.

We assessed \bertbitf outputs from the same aspects of
\emph{grammaticality} and \emph{informativeness}, as described
in~\Cref{sec:nsc_erran_uni}.

\paragraph{Grammaticality.} Out of 200 compressions, only 44 (22 \%)
were found to be grammatical; we classified the errors in the
remaining instances into groups (marked with \emph{G}
in~\Cref{fig:nsc_erran_bitf}). The first and most prevalent is the
already mentioned \emph{stitch} group which comprises around 80 \% of
all grammatical errors:
\begin{itemize}
    \item \bsp{The program has received FBS college game 2014 season .}
    \item \bsp{Tskhinvali region with Russia .}
\end{itemize}

The remaining errors are faulty compression endings (\emph{finish}):
\begin{itemize}
  \item \bsp{The fine has been described as a slap on the .}
  \item \bsp{P Chidambaram sought .}
\end{itemize}

\paragraph{Informativeness.} A similar situation was observed when
assessing the compressions' informativeness~--- only 41 (20.5 \%)
instances were considered as correct. The distribution of errors
(marked with \emph{I} in~\Cref{fig:nsc_erran_bitf}) indicates that
more than 80 \% of cases miss information by omitting important words:
\begin{itemize}
    \item \bsp{Dickinson was a .}
    \item \bsp{Wynalda is mixing .}
\end{itemize}

A smaller fraction of errors was comprised by the cases with
unresolved personal pronouns:
\begin{itemize}
    \item \bsp{He is an education .}
    \item \bsp{It would win 45 to 55 seats in Odisha .}
\end{itemize}

The remaining errors were the cases where the system compressions
changed the semantics of the input:
\begin{itemize}
    \item \textbf{Sentence}: \bsp{612 ABC Mornings intern Saskia Edwards hit the streets of Brisbane to find out what frustrates you about other people.}
    \item \textbf{System}: \bsp{Saskia frustrates people .}
\end{itemize}

More examples of \bertbitf errors are provided in~\Cref{sec:appendix_erran_bitf}.

We counted the cases in which predictions of \bertbitf had better or
worse quality, compared to \bertuni. In terms of informativeness,
\bertbitf improved 15 and worsened 78 instances; in terms of
grammaticality, 115 instances were perceived as less grammatical,
versus only 13 improved cases, which makes it clear that \bertbitf
makes many more mistakes than \bertuni, despite the higher metric
scores.

In order to verify our findings, we examined the ground-truth
compressions in more detail. Only 63 (31.5 \%) of these compressions
were both grammatical and informative. \Cref{fig:nsc_erran_tgt} shows
a visualization of the error type distribution. We provide examples of
noisy ground-truth compressions in~\Cref{sec:appendix_erran_tgt}.

The abundant errors related to the use of pronouns in the compressions
were predominantly caused by the fact that many instances contained
ground-truth compressions with unresolved pronouns; cleaning the data
would likely result in better outputs.

The \emph{stitch}, \emph{finish} and \emph{info\_miss} errors can be attributed
to the fact that many references have missing information or artifacts
remaining from the automatic procedure that was used to create these
compressions~\cite{filippova2013overcoming}. Resolving these issues
may require more elaborate strategies, beyond simple text
substitution.

\section{Discussion}

In this study we advanced the \sota for the task of sentence
compression, and achieved that by designing a simple, but effective
sequence labeling system based on the Transformer neural network
architecture. While the proposed approach achieved the highest scores
reported in the research literature, the main message of the study is
not a higher score~--- it is the idea that NLP system evaluation might
need to go beyond simple comparison of metric scores with human
judgements.

We found that a higher-scoring system can produce worse-quality
outputs. We further provided some empirical evidence that this issue
is caused by the noise in the training data. We call this finding a
\emph{discrepancy discovery}, because existent sentence compression
work does not explain our results, based on the established evaluation
practices. The research papers we analyzed present automatic and human
evaluation statistics that seem to overlook the data quality issue. Of
course, the approaches proposed so far could still produce
high-quality sentence compressions, but the absence of error analysis
plants a seed of doubt into the reader. In this work, we question not
the reported results, but the principles of the conventional
evaluation workflow.

None of the examined research papers drew attention to the quality of
the data, even though it is known that the dataset was constructed
automatically, and therefore should contain noisy examples, which
should affect the output quality of any data-driven system. Previous
work also overlooked the use of the compression ratio which seems to
be too simplistic to call it a metric that measures the compression
effectiveness. Finally, the employed sentence compression evaluation
protocols do not assume having multiple references. We did not go into
much detail about this issue, but provided an illustration at the
beginning of the paper (the \bsp{Dickson} example). The space of
possible compressions in deletion-based sentence compression is bound
by sentence length. But because the definition of importance is left
out, the candidate space is very large. The existence of only one
reference brings additional requirements for evaluation metrics to
work, and commonly used n-gram overlap metrics clearly do not satisfy
these requirements.

\section{Conclusion}
The presented results show that system output analysis is
indispensable when assessing the quality of NLP systems. It is a
well-established fact that metric scores used for system development
do not always reflect the actual quality of a system; usually this is
revealed via human evaluation experiments. However, in our case study
of automatic sentence compression we have discovered that they might
not be sufficient. Further investigation is needed to make stronger
claims; the study's findings are yet to be confirmed for other
datasets and, perhaps, tasks.

\section*{Acknowledgments}
We thank Andy Almonte, Joshua Bambrick, Minh Duc Nguyen, Vittorio
Selo, Umut Topkara and Minjie Xu for their support and insightful
comments.

\bibliography{references}
\bibliographystyle{coling}

\appendix
\section{Error Examples}\label{sec:appendix}
This section contains error examples of \bertuni and \bertbitf models,
as well as errors in the gold standard. We publicly share the
prediction and error analysis files at
\url{https://github.com/UKPLab/arxiv2021-evaluation-discrepancy-nsc}.

\paragraph{Grammaticality.} Error types:
\begin{itemize}
  \item \emph{finish}: incomplete sentence with an abrupt ending, caused by omitting the last token(s);
  \item \emph{stitch}: grammatical clauses with missing linking words, as if they are stitched together, which renders the sentence ungrammatical;
  \item \emph{rd-pron}: incorrect sentence start with a relative or demonstrative pronoun;
  \item \emph{verb-miss}: missing a verb which is essential for ensuring the grammaticality of a sentence.
\end{itemize}

\paragraph{Informativeness.} Error types:
\begin{itemize}
  \item \emph{info-miss}: missing certain information that is needed for understanding the context;
  \item \emph{p-pron}: starting with an unresolved personal pronoun, which makes it hard to understand the subject;
  \item \emph{mean-change}: omitting the context causing a change of the meaning of the sentence.
\end{itemize}

We also provide examples of alternative compressions which focus on
parts of the input sentence, which are different from those present in
the reference compressions. They are marked with an \emph{I2} label
and are listed here, together with erroneous cases, because due to
mismatches with references they lower the metric scores of the
evaluated approaches.

\subsection{\bertuni}\label{sec:appendix_erran_uni}
\subsubsection*{Grammatical Errors}
\small{
\begin{longtable}{p{0.07\textwidth} p{0.50\textwidth} p{0.33\textwidth}}
\\
\toprule
\textbf{Error} & \textbf{Sentence} & \textbf{Compression} \\ 
\midrule
finish 
 & A young cow trying to grab a cooling drink from a river in Hampshire in the hot weather had to be rescued by firefighters when it got stuck in the mud. & A cow trying to grab a drink had rescued . \\
& Japan and the US reaffirmed Monday during a meeting in Tokyo with visiting Under Secretary for Terrorism and Financial Intelligence David S Cohen their ongoing cooperation on sanctions against Iran, a Japanese government source said Tuesday. & Japan and the US reaffirmed . \\
 & 612 ABC Mornings intern Saskia Edwards hit the streets of Brisbane to find out what frustrates you about other people. & Saskia Edwards hit . \\
 & Google Helpouts, which was launched this week, is a service that allows you to pay for brief one-on-one webcam master classes with a range of experts in various fields. & Google Helpouts is . \\

\midrule
rd-pron 
& Some parents and others in Bessemer City are complaining about a YouTube video that shows young people rapping while flashing cash and a handgun in a public park. & That shows young people rapping while flashing cash and a handgun in a public park . \\
 & ``In my district, already reeling from the shutdown of our largest private employer, the highest energy costs in the country, and reduced government revenues, this shutdown, if it continues any longer, can be the final nail in our economic coffin'', said Rep. Christensen. & This shutdown can be the final nail in our economic coffin . \\
 
\midrule
stitch 
& Our obsession with Sachin Tendulkar and records has made us lose perspective to such an extent that what should have been widely condemned is being conveniently ignored. & Should have been condemned is . \\
 & Direct Link Pisegna JR. Without will amoxicillin work for a uti this sweet little boy told his best friend that he loved him. & boy told his he loved . \\
 & Visitors will find a mixture of old and new at Silver Springs State Park, which opened Tuesday near Ocala. & will find and at Silver Springs State Park opened . \\
 & Twitter took the first step toward its IPO by filing with US regulators on Thursday, September 12, 2013. & Twitter took by filing with US regulators . \\
 & Arsenio Hall lost control of his brand new Porsche Cayenne S and crashed his car on Monday night in El Lay! & Arsenio Hall lost his and crashed his car in El Lay \\
 
\midrule
verb-miss & People watch giant waves crash against the already damaged railway line and buildings at Dawlish during storms in south west England February 8, 2014. & People giant waves crash against the railway line and buildings at Dawlish . \\ 

\bottomrule
\caption{Manual error analysis results: examples of \emph{ungrammatical} outputs of \bertuni.}
\label{tab:app_err_examples_uni_gram}
\end{longtable}
}

\subsubsection*{Informativeness Errors}
\small{
  \begin{longtable}{p{0.07\textwidth} p{0.50\textwidth} p{0.33\textwidth}}
    \\
    \toprule

    \textbf{Error} & \textbf{Sentence} & \textbf{Compression} \\ 

    \midrule
    info-miss
    & Buyers should beware, even though there was a safety recall on some GM cars those vehicles are still being sold on Craigslist. & Buyers should beware . \\
    & New results show some improvement in test scores and graduation rates for local students, but experts say there's still more work to be done. & There 's still more work to be done . \\
    & Counting of postal votes have already commenced while Elections Commissioner Mahinda Deshapriya stated that he first Postal Vote result can be announced before 10PM. & He first Postal Vote result can be announced before 10PM . \\
    & When President Obama was elected in 2008 for his first term, he made a presidential decision that he would not give up his blackberry. & He would not give up his blackberry . \\
    
    \midrule
    mean-change 
    & On this week's ``Hostages'' season 1, episode 13: ``Fight or Flight,'' Ellen reveals to Duncan that she will not kill the President but will help him get what he needs, as long as he gives her something in return. & Ellen reveals she will not kill the President . \\
    & Serbia's First Deputy Prime Minister Aleksandar Vucic spoke in Germany with former German chancellor Helmut Kohl about Serbia's path towards the EU and its economic recovery/ During the talks, Vucic highlighted the important role of German investors in Serbia and voiced hope that Germany will give even stronger support to Serbia in the realisation of its European goals. & Aleksandar spoke Germany will give stronger support to Serbia . \\

    \midrule
    p-pron 
    & ``Being a five-time champion, he knows how to handle pressure. Anand generally puts a lid on his emotions.'' & He knows how to handle pressure \\
    & Chelsea manager Jose Mourinho has made light of the managerial instability under Roman Abramovich by admitting he's trying to break his own record. & He 's trying to break his own record . \\
    & Katy Perry wasn't lying when she said she had some ``beautiful news to share,'' because she is now the new face of COVERGIRL. & She had some beautiful news to share . \\

    \midrule
    I2 
    & Provisur Technologies has entered into an agreement with Scanico in which Scanico has become Provisur's global partner in commercial freezing technology. 
    & \textbf{Reference}: Scanico has become Provisur 's global partner in commercial freezing technology . \\
    &  & \textbf{System}: Provisur Technologies has entered into an agreement with Scanico . \\
    & Davina McCall has undergone medical tests after fears she may be suffering from hypothermia after battling severe weather in a Sport Relief challenge.
    & \textbf{Reference}: she may be suffering from hypothermia after battling in a Sport Relief challenge . \\
    &  & \textbf{System}: Davina McCall has undergone medical tests . \\ 
    \bottomrule
    \caption{Manual error analysis results: examples of
      \emph{uninformative} outputs of \bertuni. We also show examples of
      alternative compressions (\emph{I2}) which deviate from ground-truth
      compressions, but cannot be considered as errors.}
    \label{tab:app_err_examples_uni_info}
  \end{longtable}
}

\subsection{\bertbitf}\label{sec:appendix_erran_bitf}
\subsubsection*{Grammatical Errors}
\small{
\begin{longtable}{p{0.07\textwidth} p{0.50\textwidth} p{0.33\textwidth}}
\\
\toprule
\textbf{Error} & \textbf{Sentence} & \textbf{Compression} \\ 
\midrule
finish 
& A UKIP candidate who is standing for election in Enfield Town has defended a tweet in which he said a black comedian ``should emigrate to a black country.'' & UKIP candidate has defended a tweet he said a black comedian should \\
 & Wynalda, who was introduced as the Silverbacks' new manager on Tuesday, is mixing a bit of Europe with a bit of Mexico with a bit of Silicon Valley in an approach that will eliminate the head-coaching position. & Wynalda is mixing . \\
 & Cell C has announced new pre-paid and contract packages that offer unlimited calls to any network. & Cell C has announced . \\
 & Radnor police received a report Sept. 3 from a cadet at Valley Forge Military Academy that another cadet struck him in the face. & Radnor police received a report another . \\
 & Diego Forlan scored directly from a corner this weekend to help Internacional to a 3-2 win over Fluminense. & Diego Forlan scored to help Internacional to . \\
 & This 1930s-built four bedroom detached seafront home in Worthing is immaculately presented and has been expertly modernised. & This seafront home is . \\

\midrule
stitch 
& Buyers should beware, even though there was a safety recall on some GM cars those vehicles are still being sold on Craigslist. & Buyers should beware are being sold on Craigslist . \\
 & Davina McCall has undergone medical tests after fears she may be suffering from hypothermia after battling severe weather in a Sport Relief challenge. & Davina McCall has undergone tests be suffering from hypothermia . \\
 & New results show some improvement in test scores and graduation rates for local students, but experts say there's still more work to be done. & Show some improvement in test scores graduation rates students but there . \\
 & POLICE are looking for witnesses after a car was hit by a van which failed to stop on Friday, January 7. & POLICE a . \\ 
 & Our obsession with Sachin Tendulkar and records has made us lose perspective to such an extent that what should have been widely condemned is being conveniently ignored. & With Sachin Tendulkar has made lose perspective what .\\
 & Watford have this evening announced the signing of experienced defender Lucas Neill, who joins for the remainder of the season subject to international clearance. & Watford announced the signing Lucas season . \\
 \bottomrule
\caption{Manual error analysis results: examples of \emph{ungrammatical} outputs of \bertbitf}.
\label{tab:app_err_examples_bitf_gram}
\end{longtable}
}

\subsubsection*{Informativeness Errors}
\small{
\begin{longtable}{p{0.07\textwidth} p{0.50\textwidth} p{0.33\textwidth}}
\\
\toprule
\textbf{Error} & \textbf{Sentence} & \textbf{Compression} \\ 
\midrule
info-miss 
 & ``Former President Mandela is still in a critical condition in hospital but shows sustained improvement,'' President Jacob Zuma said in a statement. & Mandela is still in a critical condition shows improvement Jacob . \\
 & This 1930s-built four bedroom detached seafront home in Worthing is immaculately presented and has been expertly modernised. & This seafront home is . \\
 & ASI's additional director general BR Mani said he was hopeful of Nalanda making it to the list, claiming that Nalanda was an important centre of art and culture even before the university came into being. & He Nalanda was . \\
 & Tata Martino explained that after two months in Barcelona, he is starting to understand how things work at the club, assuring that there seems to be a need to create a weekly crisis regarding the team's style of play. & He crisis . \\

\midrule
mean-change 
& Provisur Technologies has entered into an agreement with Scanico in which Scanico has become Provisur's global partner in commercial freezing technology. & Provisur has become Provisur 's partner in commercial freezing technology . \\
 & TV presenter Katie Piper shows off her baby daughter for the first time and said motherhood has given her ``true fulfillment''. & Piper shows off her baby daughter motherhood . \\
 & AFL powerbrokers yesterday met Collingwood officials over convicted criminal Marley Williams, who faces a possible club ban or even a league suspension. & Powerbrokers met ban suspension . \\
 & British rapper Professor Green was arrested on suspicion of drink driving on Sunday after he called police to report a robbery. & Professor Green was arrested on suspicion of report a robbery . \\

\midrule
p-pron 
 & ``Being a five-time champion, he knows how to handle pressure. Anand generally puts a lid on his emotions.'' & He puts a lid on his emotions \\
 & PersonalWeb® today announced that it has reached a confidential settlement in its lawsuit with NEXSAN®, to which Nexsan has licensed the Truename patents from PersonalWeb. & It has reached patents PersonalWeb . \\
 & Britney Spears has said she hopes her latest album Britney Jean will inspire people and she wants to 'project positive energy out into the world. & She hopes her album Britney will inspire . \\

\midrule
I2
& On this week's ``Hostages'' season 1, episode 13: ``Fight or Flight,'' Ellen reveals to Duncan that she will not kill the President but will help him get what he needs, as long as he gives her something in return. 
    & \textbf{Reference}: On this week 's Hostages season 1 episode 13 Fight or Flight Ellen reveals . \\
 &  & \textbf{System}: Episode Fight or Flight Ellen reveals she will not kill the President . \\
 & Blustery winds arrived in Gwinnett on Wednesday and brought with them lower temperatures that caused the National Weather Service to issue a freeze warning for the area. 
    & \textbf{Reference}: that caused the National Weather Service to issue a freeze warning . \\
 &  & \textbf{System}: Blustery winds arrived issue a freeze warning . \\
 & BJP and JD today welcomed the five-year jail term handed down to RJD chief Lalu Prasad in the fodder scam case, saying it would send out a message that the law will catch up with the corrupt, however influential they might be. 
    & \textbf{Reference}: The law will catch up with the corrupt influential . \\
 &  & \textbf{System}: BJP law will catch up with the corrupt . \\
\bottomrule
\caption{Manual error analysis results: examples of \emph{uninformative} outputs of \bertbitf. We also show examples of alternative compressions (\emph{I2}) which deviate from ground-truth compressions, but cannot be considered as errors.}
\label{tab:app_err_examples_bitf_info}
\end{longtable}
}

\subsection{Ground Truth}\label{sec:appendix_erran_tgt}
\subsubsection*{Grammatical Errors}
\small{
  \begin{longtable}{p{0.07\textwidth} p{0.50\textwidth} p{0.33\textwidth}}
    \\
    \toprule
    \textbf{Error} & \textbf{Sentence} & \textbf{Compression} \\ 
    \midrule
    finish 
    & Police investigating the unexplained death of a man in Taupo say his van appears to have broken down. & Police investigating the unexplained death say . \\
    & Akkineni Nageswara Rao was one of the Indian cinema's stalwarts, who will be remembered for his rich contribution. & Akkineni Nageswara Rao was one . \\
    & Mortgage fees are going up so where does Pa. & Where does Pa . \\
    & Coffee chain Starbucks has said guns are no longer ``welcome'' in its US cafes, although it has stopped short of an outright ban. & Starbucks has said guns are . \\

    \midrule
    rd-pron
    & Way back in May 2011, Google filed a patent application for eye tracking technology, which would allow it to charge advertisers on a 'pay per gaze' basis. & Which would allow it to charge advertisers on a pay per gaze basis . \\
    & POLICE are looking for witnesses after a car was hit by a van which failed to stop on Friday, January 7. & Which failed to stop . \\
    & Tomorrow South Africa will celebrate the centenary of the Union Buildings in Pretoria which have recently been declared a national heritage site by the South African Heritage Resources Agency. & Which have been declared a national heritage site . \\
    & In a press release, Patrick said Goldstein will be replaced by Rachel Kaprielian, who is currently the state's registrar of motor vehicles. & Who is the state 's registrar .\\

    \midrule
    stitch 
    & Iran wants to end the stand-off with global powers over its nuclear programme swiftly, but will not sacrifice its rights or interests for the sake of a solution, President Hassan Rouhani said on Friday. & Iran wants but will not sacrifice its rights Hassan Rouhani said . \\
    & Maggie Rose sheds her innocence in her brand new music video for ``Looking Back Now.'' & Maggie Rose sheds for Looking Back Now \\
    & The Muskingum University chapter of Omicron Delta Kappa has made a donation of more than \$600 to the New Concord Food Pantry in an effort to give back to the community. & The Muskingum University chapter of Omicron Delta Kappa has made in an effort to give back to the community . \\
    & Dolly Bindra filed a case on an unknown person for having threatened her at gun point today in Oshiwara, Mumbai. & Dolly Bindra filed for having threatened her at gun point in Oshiwara Mumbai . \\
    & North Korean leader Kim Jong-un has met with the top military leaders and warned them of a grave situation and threatened a new nuclear test. & Kim Jong-un has met and warned of a grave situation and threatened a nuclear test . \\

    \bottomrule
    \caption{Manual error analysis results: examples of
      \emph{grammatical errors} in ground-truth compressions, sampled
      from 200 development set instances with lowest \bertuni F1
      scores).}
    \label{tab:app_err_examples_tgt_gram}
  \end{longtable}
}

\subsubsection*{Informativeness Errors}
\small{
  \begin{longtable}{p{0.07\textwidth} p{0.50\textwidth} p{0.33\textwidth}}
    \\
    \toprule
    \textbf{Error} & \textbf{Sentence} & \textbf{Compression} \\ 
    \midrule
    info-miss
    & Some parents and others in Bessemer City are complaining about a YouTube video that shows young people rapping while flashing cash and a handgun in a public park. & Some parents in Bessemer City are complaining about a video . \\
    & Tata Martino explained that after two months in Barcelona, he is starting to understand how things work at the club, assuring that there seems to be a need to create a weekly crisis regarding the team's style of play. & There seems to be a need to create a weekly crisis .\\
    & Nothing is ever left behind in a BREACHED performance as the loud rocking, heavy amp cranky band announce Toronto show dates since performing last October at Indie Week. & Band announce Toronto show dates \\
    & Prime Minister Kevin Rudd has missed the deadline for an August 24 election, with his deputy saying ``people should just chill out'' about the election date. & People should chill out about the election date . \\
    \midrule
    p-pron 
    & Davina McCall has undergone medical tests after fears she may be suffering from hypothermia after battling severe weather in a Sport Relief challenge. & She may be suffering from hypothermia after battling in a Sport Relief challenge . \\
    & TV presenter Nick Knowles has been the recipient of some unexpected abuse as a result of an announcement that he will not be present at the birth of his child. & He will not be present at the birth . \\
    & England fast bowler James Anderson does not feel sorry for Australia and has said his team wants to win the Ashes 5-0. & His team wants to win the Ashes 5 0 . \\
    & If he decides to run for president, New Jersey Gov. Chris Christie will need to push back against the inevitable pressure that he will encounter to move to the right. & He will encounter to move to the right . \\
    & Armaan will be taken for a medical examination and post that he will be presented in the court today. & He will be presented in the court . \\
    \bottomrule
    \caption{Manual error analysis results: examples of \emph{informativeness} errors in
      ground-truth compressions, sampled from 200 development set
      instances with lowest \bertuni F1 scores).}
    \label{tab:app_err_examples_tgt_info}
  \end{longtable}
}

\end{document}